\documentclass[runningheads]{llncs}
\usepackage{graphicx}
\usepackage{epsfig, multirow, multicol}
\usepackage{amssymb, booktabs}
\usepackage{amsmath}
\usepackage{cite}
\begin{document}
\title{DPFNet: A Dual-branch Dilated Network with Phase-aware Fourier Convolution for Low-light Image Enhancement %\thanks{Supported by organization x.}
}
\titlerunning{Dual-branch Dilated Network with Phase-aware Fourier Convolution}
\author{Yunliang Zhuang\inst{1} \and
Zhuoran Zheng \inst{2} \and
Yuang Zhang \inst{1} \and
Lei Lyu\inst{1} \and
Chen Lyu\inst{1}}
%a 
\authorrunning{Zhuang et al.}
\institute{School of Information Science and Engineering, Shandong Normal University, Jinan, 250014, China \\
\email{lvchen@sdnu.edu.cn} \and
School of Computer Science and Engineering, Nanjing University of Science and Technology, Nanjing, 210014, China}

\maketitle              % typeset the header of the contribution
\begin{abstract}
Low-light image enhancement is a classical computer vision problem aiming to recover normal-exposure images from low-light images. However, convolutional neural networks commonly used in this field are good at sampling low-frequency local structural features in the spatial domain, which leads to unclear texture details of the reconstructed images. To alleviate this problem, we propose a novel module using the Fourier coefficients, which can recover high-quality texture details under the constraint of semantics in the frequency phase and supplement the spatial domain. In addition, we design a simple and efficient module for the image spatial domain using dilated convolutions with different receptive fields to alleviate the loss of detail caused by frequent downsampling. We integrate the above parts into an end-to-end dual branch network and design a novel loss committee and an adaptive fusion module to guide the network to flexibly combine spatial and frequency domain features to generate more pleasing visual effects. Finally, we evaluate the proposed network on public benchmarks. Extensive experimental results show that our method outperforms many existing state-of-the-art ones, showing outstanding performance and potential.
We release our code at https://github.com/Zhuangyunliang/DPFNet.

%we propose a Phase-aware Fourier Convolution Module to recovered in the frequency domain to complement the spatial domain so that the recovered image contains more high-frequency details while ensuring effective utilization of the spatial domain of the image. However, the existing frequency domain-based work often ignores the phase component when processing the frequency domain information, and this processing results in the loss of image edge. %To this end, we design a frequency domain subnetwork based on complex convolution to process the amplitude and phase components to retain the more detailed edge. In addition, we propose a novel loss committee and an adaptive fusion module to guide the network to flexibly combine spatial and frequency domain features to generate more pleasing visual effects. We evaluate the proposed network on several public benchmark datasets. Extensive experimental results show that our method outperforms many of the existing state-of-the-art ones, showing great effectiveness and potential.

\end{abstract}
\section{Introduction}
\label{sec:intro}
Images as a modality with a large number of discrete units can convey a wealth of information. However, due to unavoidable environmental or technical limitations (such as inadequate lighting and limited exposure time), images taken in low-light conditions do not effectively convey real-world color and texture information.
Meanwhile, low-light images pose challenges for high-level computer vision tasks such as semantic segmentation, object detection, and image classification \cite{wang2022sfnet, gnanasambandam2020image, li2021deep}. 
Therefore, enhancing the luminance and texture of low-light images for high-level vision tasks is a non-trivial problem. 

Currently, most researchers propose various methods that can effectively improve the subjective and objective quality of low-light images.
Traditional low-light image enhancement methods \cite{ying2017bio, ren2018lecarm, guo2016lime} use nonlinear mapping of pixel values from low-light images to achieve enhanced images.
However, the carefully chosen parameters and optimization process of this class of methods require much knowledge and experience from the user and cannot be generalized to real-world scenes.
Recently, learning-based methods \cite{wang2021low} have shown great potential and have received widespread attention for their better effectiveness, robustness, and generalizability than traditional methods.
For instance, Lore et~al. \cite{llnet} first applied deep neural networks to low-light image enhancement tasks and proposed an algorithm based on an autoencoder structure.
Liang et~al. \cite{liang2021swinir} proposed a transformer-based image restoration method.

Despite the significant performance improvements of deep learning-based methods, 
several significant challenges remain for low-light image enhancement tasks:
%significant challenges remain for low-light image enhancement tasks, as follows:
%
\textbf{1)}
%Current methods rarely capture the gap between the reconstructed image and the ground truth from the frequency domain perspective, which is also critical for low-light image enhancement. Specifically, the frequency domain of the image can restore high-frequency texture details, making up for the loss of information in the spatial domain \cite{93}.
The frequency domain information of the image helps to reconstruct high-frequency texture details, making up for the loss of information in the spatial domain \cite{93}. However, due to the limitation of real-valued convolution kernels, existing work \cite{suvorov2022resolution} usually omit or separately processes phase information, making the reconstruction process lack semantic guidance in phase.
\textbf{2)} Most of the above methods use stacked convolution kernels and up/down sampling to learn the features of global and different receptive fields. However, such schemes tend to lead to loss of texture details, which can seriously affect the result of the reconstructed images \cite{lim2020dslr}.
\textbf{3)} Transformer-based method shows strong competitiveness, proving the effectiveness of long-term dependence on the low-level image processing tasks. However, due to the characteristics of the self-attention mechanism, such methods usually depend on expensive hardware resources and a large amount of data samples \cite{wang2022uformer, zamir2022restormer}.

To alleviate the above challenges, we propose a dual-branch dilated Network with phase-aware Fourier convolutions, named DPFNet. DPFNet can enhance low-light images in the frequency and spatial domains, respectively.
In particular, we design a phase-aware Fourier convolution module to provide additional information for spatial domain reconstruction by exploiting the frequency domain characteristics after the fast Fourier transform (FFT).
The module achieves the perception of semantic information in the phase by sharing the weights of the memory to combine the magnitudes for better performance dynamically. Due to the characteristics of the FFT, it can cover the receptive field to the entire image space at an early stage and efficiently perceive the global difference between low-light and normal-exposure images.
In addition, we adopt a dilated convolution module with multi-level receptive fields to perform the enhancement on the spatial domain. This module can reduce the loss of image detail information caused by successive downsampling and economize the computational effort. Finally, we introduce a fusion module to fuse the enhancement results from the spatial and frequency domains and design a Fourier loss added to the loss committee to balance the different domains.
Extensive experiments demonstrate that our proposed DPFNet outperforms state-of-the-art methods on public datasets.

In summary, our main contributions are threefold:
\begin{itemize}
%	\item We present a new hybrid representation guidance network with two augmented branches to improve the spatial and frequency domain information of the image. The enhancement performance for low-light images benefits from the combination of the two augmented branches within one joint framework. 
    % \item We propose a new low-light enhancement network architecture that performs targeted enhancement of features in different information domains of the image and dynamically fuses the features extracted from different information domains.
    
    \item
    %We design a phase-aware Fourier convolution module to reconstruct high-frequency texture details in the Fourier space of the low-light images to complement the spatial domain.
    %
    We design a phase-aware Fourier convolution module that reconstructs high-frequency texture details guided by phase semantic information in image Fourier space to complement the spatial domain.
    %In particular, we propose a novel frequency-domain sub-network to construct the magnitude and phase information at different frequencies in the image frequency domain to complement the loss in the spatial domain.
    %In particular, we use FFT to transform the image into frequency-domain space and build specific frequency domain sub-networks to construct the magnitude and phase information at different frequencies.

% 	\item In the frequency domain branch, we proposed a novel FDN to decompose the image into different frequency components, making full use of the magnitude and phase information of the Fourier coefficients to enhance the image from the frequency domain perspective.
    % \item Our proposed HRGNet uses the FFT to decompose the image into different frequency components, which can make full use of the real and imaginary part information of the Fourier coefficients to enhance the image from the frequency domain perspective.
    \item We integrate spatial and frequency domain enhancements into an end-to-end dual-branch network. The network is capable of integrating low- and high-frequency information while capturing local and global interactions.
    
    \item We propose a Fourier loss added to the loss committee to balance features in the spatial and frequency domains to restore natural images.
	
	%\item Our network is compared with many competing methods through comprehensive experiments. The results show that our approach uses only the base module to establish a new state-of-the-art low-light image enhancement performance.
\end{itemize}

% The rest of this paper is organized as follows: Section \ref{sec:rela} reviews previous studies relevant to our work; Section \ref{sec:method} describes our methodological framework and elaborates its details; Section \ref{sec:exp} compares the performance of the proposed method with other research methods performance, conducts ablation experiments on our method. Finally, our work is summarized in Section \ref{sec:con}.
%and conclude with a summary of our work.

%-------------------------------------related work-------------------------------------
\section{Related Work}
\label{sec:rela}
\subsection{Deep Low-light Image Enhancement}
With incentives from the good results of deep learning in various tasks, many researchers have started investigating low-light image enhancement algorithms \cite{Retinex, lim2020dslr, wu2022uretinex} based on deep learning. Lore et~al. \cite{llnet} proposed an autoencoder deep learning network for low-light image enhancement. Zhang et~al. \cite{KinD} developed three sub-networks for layer decomposition, reflection recovery, and illumination adjustment, called KinD. Wang et~al. \cite{wang2021low} used a normalized flow model to establish a mapping between low-light and normal-light images. This allows for determining the correct conditional distribution of the enhanced images. However, these methods focus only on the spatial domain when enhancing low-light images and ignore information in the frequency domain. We propose a PFM that acts in the image frequency domain to recover the high-frequency texture details of normally-exposed images. In addition, such methods often use continuous downsampling to obtain large ﬁelds of perception, which leads to a loss of detailed information. We use dilated convolutions with different dilation rates to obtain information from different receptive fields, thereby mitigating the downsampling damage.
% \vspace{-2 mm}
\subsection{Applications of Fourier Transform}
In recent years, the Fourier domain of images is often applied to accomplish different tasks with favorable results, which provides a new perspective for computer vision tasks based on images. Rao et al. \cite{rao2021global} proposed that GFNet learns spatial long-term dependencies from the frequency domain and shows power competitiveness in image classification tasks. Mao et al. \cite{mao2021deep} designed a plug-and-play Fourier module to enhance the details of the deblurring algorithm. Wang et al. \cite{suvorov2022resolution} designed a fast Fourier convolution to improve the perceptual quality and parameter efficiency of mask inpainting networks. However, most methods directly use the amplitude component of the Fourier coefficients or simply relate phase and amplitude, ignoring the semantic guidance of phase. Thus, we carefully design the Fourier convolution module to strengthen the connection between the two through associative memory with conjugate symmetry constraint on the spectral-domain weights.
\begin{figure}[t]
	\centering
		\includegraphics[width=\textwidth]{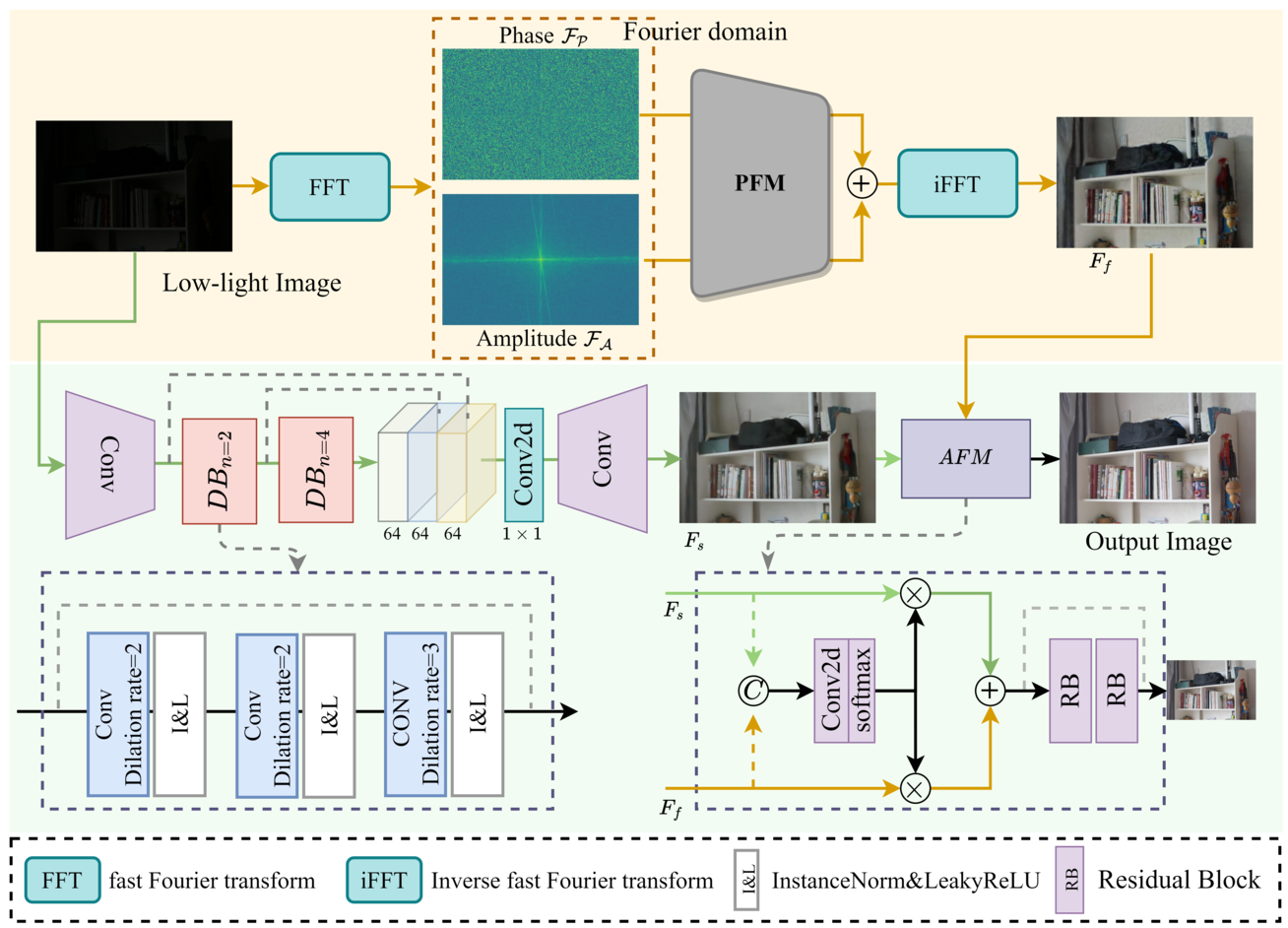}
	  \caption{The overall architecture of our proposed DPFNet.
	  %The architecture of the proposed single image low-light enhancement network consists of three parts. The first part is a sub-network based on spatial domain information, which extracts the higher-order spatial domain characteristics with the help of multi-scale dilation convolutional layers; the second part is a sub-network based on frequency domain information, which captures the texture of the image. The last part is a feature adaptive fusion module (AFM), which dynamically allocates weights on the products of the dual branch.
	  }
	  \label{fig:fig1}
	  \vspace{-4mm}
\end{figure}

%----------------------------------method----------------------------------
\section{Methodology}
\label{sec:method}
This section describes our proposed DPFNet. Inspired by \cite{koh2022bnudc, yu2021two}, we adopt a network structure with two independent branches to avoid the noise introduced by the mutual interference between the spatial and frequency domains during processing. DPFNet is mainly composed of two feature enhancement streams and a feature fusion. As shown in Figure \ref{fig:fig1}, given the low-light images $I_{low}$ as input, our network uses the phase-aware Fourier convolution module (PFM) to generate feature maps in the Fourier domain that contain more global and high-frequency details (upper stream). Then, the multi-level dilated convolution module (MDCM) aggregates the local contextual and content features under different receptive fields in the spatial domain (bottom stream). Finally, the adaptive fusion module (AFM) fuses the spatial and frequency domain features to avoid over-enhancement artifacts and reconstruct more natural, high-quality images. We describe the design of each part of DPFNet in detail below.

% ------------- network-----------------
\begin{figure}[t]
	\centering
		\includegraphics[width=0.96\textwidth]{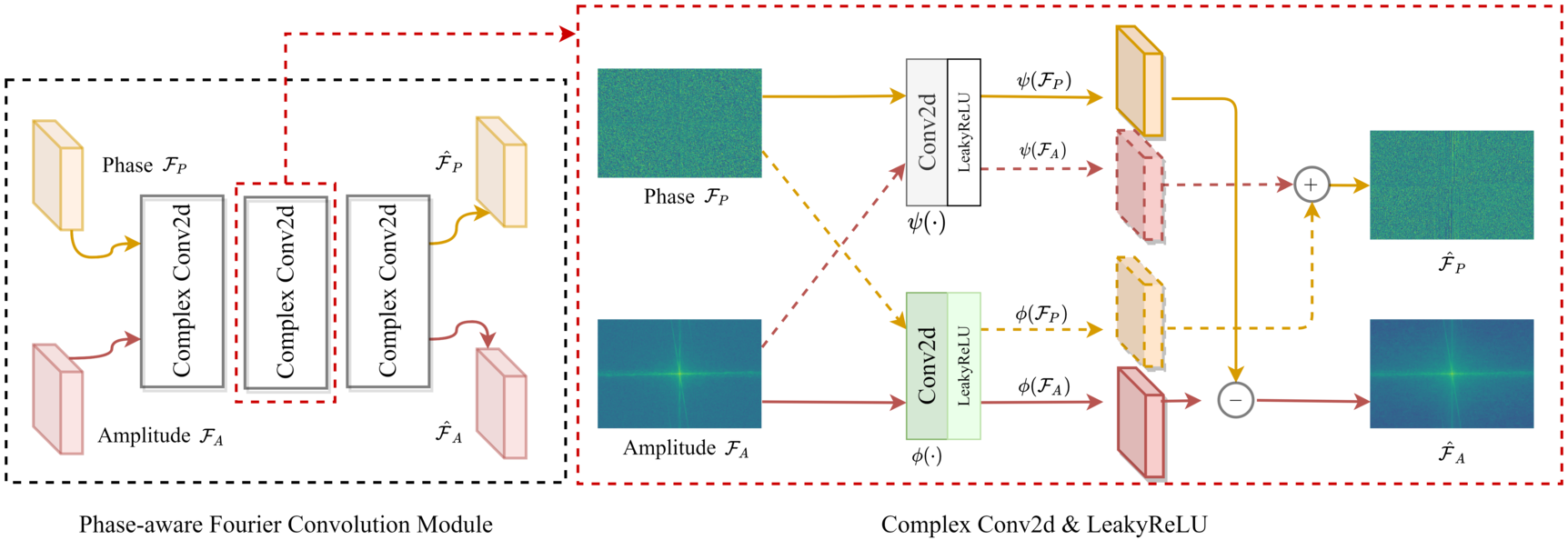}
	  \caption{The structure of our proposed Phase-aware Fourier Convolution Module.
	  %The architecture of the proposed single image low-light enhancement network consists of three parts. The first part is a sub-network based on spatial domain information, which extracts the higher-order spatial domain characteristics with the help of multi-scale dilation convolutional layers; the second part is a sub-network based on frequency domain information, which captures the texture of the image. The last part is a feature adaptive fusion module (AFM), which dynamically allocates weights on the products of the dual branch.
	  }
	  \label{fig:fig2}
	  \vspace{-4mm}
\end{figure}

\subsection{Phase-aware Fourier Convolution Module}
The PFM aims to restore the high-frequency texture details with the assistance of frequency domain information to supplement the spatial domain. The input low-light image $I_{low}$ is converted into the frequency domain by the FFT. It can be expressed as follows:
\begin{equation}
	\mathcal{F}(u, v)=\sum_{x, y} I_{low}(x, y) e^{-j 2 \pi\left(\frac{u}{M}x+\frac{v}{N}y\right)},  
\end{equation}
where $I_{low}(x, y)$ and $\mathcal{F}(u, v)$ are the spatial coordinates of the pixel and spectrum of the frequency domain, the height and width of the images are denoted $(M, N)$. Obviously, the spectrum of arbitrary frequency contains the global information under the spatial position.

We decompose the complex-valued form of the spectrum into the amplitude $\mathcal{F}_A$ that contains the color texture and the phase $\mathcal{F}_P$ that contains the semantic structure, and feed them into the PFM. To reduce the computational cost, we stack only three blocks with the same structure in the PFM, as shown in Figure \ref{fig:fig2}. The block contains three complex convolutions with the kernel being $3\times3$ and the LeakyReLU activation function. The complex convolution can enhance the phase semantic guidance \cite{hu2020dccrn}, which is achieved by two different real-valued convolution operations, where the parameters of the filters of the real-valued convolution are shared. The operation process is formulated as:
\begin{equation}
\begin{aligned}
    &\mathcal{F}_P  \gets   \phi(\mathcal{F}_P) + \psi(\mathcal{F}_A) \\
	&\mathcal{F}_A  \gets   \phi(\mathcal{F}_A) - \psi(\mathcal{F}_P),
\end{aligned}
\end{equation}
where the $\phi(\cdot)$ and $\psi(\cdot)$ are two uncorrelated convolutions. We finally use the inverse fast Fourier transform (iFFT) to reconstruct the enhanced spectrum of the frequency into spatial domain $F_f$, as follows:
\begin{equation}
     F_f = iFFT(\mathcal{F}_A + j \mathcal{F}_P),
\end{equation}
where $j$ is the unit of the complex-valued imaginary part.

\subsection{Multi-level Dilated Convolution Module}
This section gives more details about this MDCM with dilated convolution blocks (DB). As shown in Figure \ref{fig:fig1}, the low-light image $I_{low}$ obtains the feature $F_{local}$ of the local receptive field through 3-layer convolutional neural networks. Then, $F_{local}$ is sequentially fed into multiple DBs, and multi-level features under a specific receptive field are obtained. We stack features of different receptive fields in the channel dimension ($cat$) and aggregate them through non-overlapped convolutional layers with the kernel size being 1 $\times$ 1. This process is formulated as:
\begin{equation}
    	F_s = \phi(cat[DB_n(DB_{n-2}(F_{local})), DB_{n-2}(F_{local}), F_{local}]) \oplus F_{low},
\end{equation}
where $n$ represents the dilation rate of the dilated convolution in DB. Inspired by \cite{liu2021inception}, we add a convolution with the dilation rate $=n+1$ into each DB to alleviate grid artifacts.

\subsection{Adaptive Fusion Module}
To blend effectively the reconstructed high-quality features to preserve the regions with good visibility, we concatenate the extracted features $F_f$ of the frequency stream and the features $F_s$ of the spatial stream in the channel dimension and feed them to a convolutional layer to learn two importance indexes $w_f, w_s$, which can be expressed as:
\begin{equation}
    	w_{f}, w_{s} = softmax(Conv(cat[F_f, F_s])),
\end{equation}
where $Conv(\cdot)$ is the convolution layer with two kernels being $3\times3$.
In the importance indexes, each pixel is assigned a corresponding weight, which is multiplied by the corresponding feature element by element to reconstruct the outputs. The recovery process can be defined as follows:
\begin{equation}
		I_{norm} = RB_{\times 2}(F_{f} \otimes w_{f} \oplus F_{s} \otimes w_{s}),
\end{equation}
where the $\otimes$ denoted element multiplication and the $\oplus$ represents element addition. 
Note that the outputs are passed through two residual blocks (RB) to generate the final result $I_{norm}$ to capture subtle changes in image enhancement so that the enhanced image appears natural.
% %-----------------AFM------------------------------
% \begin{figure}[t]
% 	\centering
% 	\includegraphics[width=0.48\textwidth]{imgs/AFM.pdf}
% 	\caption{The proposed Adaptive Fusion Module (AFM), where yellow images represent feature maps based on spatial domain information, and green images represent feature maps based on frequency domain information. It is worth noting that after the additive operator fusion, a layer of convolution is passed in order that the fused information will not show ringing or artifacts.}
% 	\label{fig:fig3}
% \end{figure}

\subsection{Loss Function}
We adopted a loss committee consisting of three parts, each with specific capabilities.

\noindent\textbf{SSIM Loss.} Since the degradation of low-light images is related to many different factors, using $\mathnormal{l}_1$ or $\mathnormal{l}_2$ as the loss function leads to different degrees of distortion and does not give the most desirable results. In contrast, we adopt the SSIM loss to integrally evaluate differences such as luminance, contrast, and structure. Specifically, the $\mathcal{L}_{s}$ is defined as:
\begin{equation}
	\mathcal{L}_{s} = 1 - \text{SSIM}(I_{norm}, I_{gt}),
\end{equation}
where $I_{norm}, I_{gt}$ represent enhanced images and ground truth, and $SSIM(\cdot)$ denotes the SSIM \cite{zhao2016loss} operator.

\noindent\textbf{Fourier Loss.}
To improve the sensitivity of our proposed model to frequency feature, we propose a Fourier Loss ($\mathcal{L}_{f}$) based on frequency domain space to guide the model to reconstruct high-frequency detail. The $\mathcal{L}_{f}$ can be considered as a weighted average of the frequency distance between the ground truth and the enhanced image, which is
\begin{equation}
    \mathcal{L}_{f} = \frac{1}{N}\sum_{i=0}^{N}\|cat[I_{gt}^A, I_{gt}^P] - cat[I_{norm}^A, I_{norm}^P]\|^2
    %\frac{1}{N}\sum_{i=0}^{N}(\| \Re_{R_i} - \Re_{\hat{R_i}}\|+\|\Re_{L_i} - \Re_{\hat{L_i}}\| + \|\Im_{R_i} - \Im_{\hat{R_i}}\|+\|\Im_{L_i} - \Im_{\hat{L_i}}\|),
\end{equation}
where $I_{gt}^A, I_{gt}^P$ represents the amplitude and phase components of the ground truth images by the FFT transform, which are concatenated together in the channel dimension, jointly minimizing the gap between the ground truth and the enhanced image.

\noindent\textbf{Perceptual Loss.} 
%Although pixel-level loss functions provide powerful guidance, they ignore some semantic features that are difficult to measure. 
We introduce Perceptual loss ($\mathcal{L}_{p}$) \cite{johnson2016perceptual} as a perceptual measure to exploit image semantic information and improve the visual quality of enhanced images. Specifically, we use Euclidean distance to calculate the difference between feature maps, and the formula is defined as follows:
\begin{equation}
	\mathcal{L}_{p} = \frac{1}{WHC}{\| \phi(I_{gt}) - \phi(I_{norm}) \|}^2,
\end{equation}
where $W$, $H$ and $C$ denote the three dimensions of the image, respectively, and the pre-trained VGG network \cite{simonyan2014very} is denoted as $\phi(\cdot)$. $\mathcal{L}_{p}$ balances the guiding role of $\mathcal{L}_{s}$ and $\mathcal{L}_{s}$, ensuring the stability of our network training process.

\noindent\textbf{Total Loss.} We use the loss committee to construct the total loss function, which is
\begin{equation}
	\mathcal{L} = \mathcal{L}_{s} + \lambda_a \mathcal{L}_{f} + \lambda_b \mathcal{L}_{p},
\end{equation}
where $\lambda_a$ and $\lambda_b$ are trade-off weight hyperparameters. Extensive experiments demonstrate that the best results of our network are obtained when $\lambda_a$ = 1.0 and $\lambda_b$ = 0.2.
%In our experiments, we empirically set the trade-off weights of each component is 1. In addition, our model is trained end-to-end with total loss until convergence.
\begin{figure}[ht]
	\begin{minipage}[b]{0.161\linewidth}
		\centering
		\centerline{\epsfig{figure=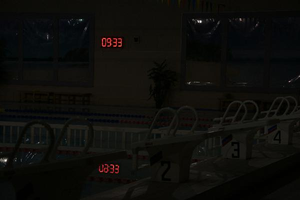,  width=\textwidth}}
		\vspace{0.1 mm}
		\centerline{\epsfig{figure=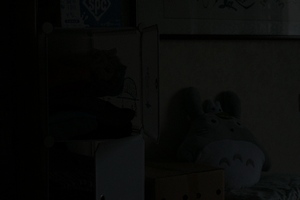,  width=\textwidth}}
		\vspace{0.1 mm}
		\centerline{\epsfig{figure=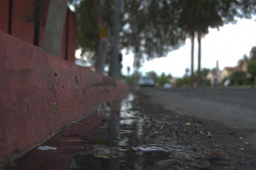,  width=\textwidth}}
		\vspace{0.1 mm}
		\centerline{\epsfig{figure=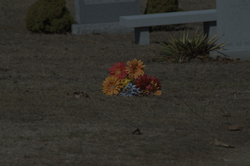,  width=\textwidth}}
		\centering{Input}
	\end{minipage}
	\begin{minipage}[b]{0.161\linewidth}
		\centering
		\centerline{\epsfig{figure=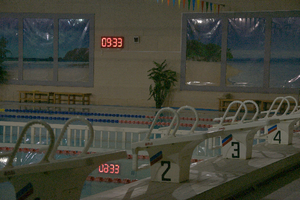,  width=\textwidth}}
		\vspace{0.1 mm}
		\centerline{\epsfig{figure=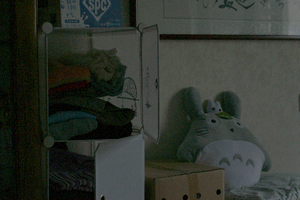,  width=\textwidth}}
		\vspace{0.1 mm}
		\centerline{\epsfig{figure=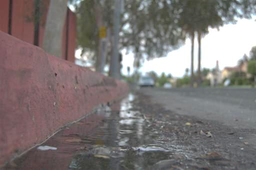,  width=\textwidth}}
		\vspace{0.1 mm}
		\centerline{\epsfig{figure=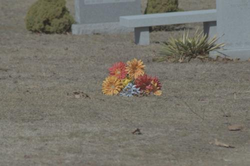,  width=\textwidth}}
		\centering{BIMEF}
	\end{minipage}
	\begin{minipage}[b]{0.161\linewidth}
		\centering
		\centerline{\epsfig{figure=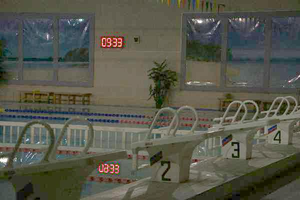,  width=\textwidth}}
		\vspace{0.1 mm}
		\centerline{\epsfig{figure=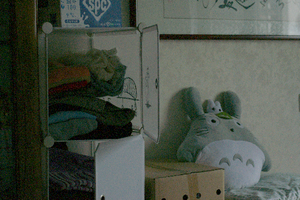,  width=\textwidth}}
		\vspace{0.1 mm}
		\centerline{\epsfig{figure=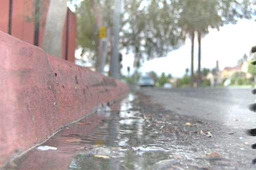,  width=\textwidth}}
		\vspace{0.1 mm}
		\centerline{\epsfig{figure=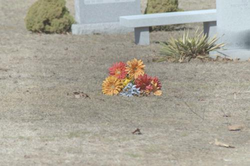,  width=\textwidth}}
		\centering{CRM}
	\end{minipage}
	\begin{minipage}[b]{0.161\linewidth}
		\centering
		\centerline{\epsfig{figure=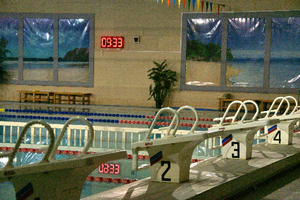,  width=\textwidth}}
		\vspace{0.1 mm}
		\centerline{\epsfig{figure=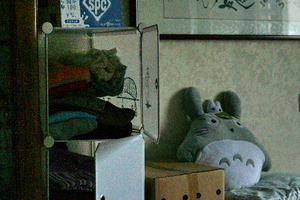,  width=\textwidth}}
		\vspace{0.1 mm}
		\centerline{\epsfig{figure=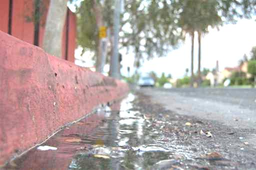,  width=\textwidth}}
		\vspace{0.1 mm}
		\centerline{\epsfig{figure=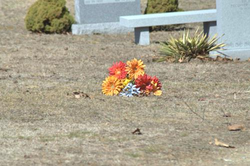,  width=\textwidth}}
		\centering{LIME}
	\end{minipage}
	\begin{minipage}[b]{0.161\linewidth}
		\centering
		\centerline{\epsfig{figure=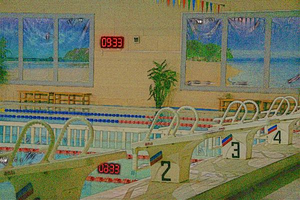,  width=\textwidth}}
		\vspace{0.1 mm}
		\centerline{\epsfig{figure=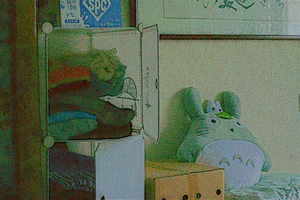,  width=\textwidth}}
		\vspace{0.1 mm}
		\centerline{\epsfig{figure=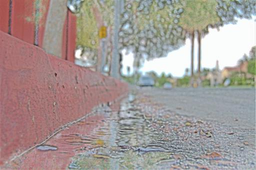,  width=\textwidth}}
		\vspace{0.1 mm}
		\centerline{\epsfig{figure=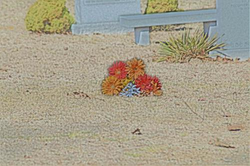,  width=\textwidth}}
		\centering{RetinexNet}
	\end{minipage}
	\begin{minipage}[b]{0.161\linewidth}
		\centering
		\centerline{\epsfig{figure=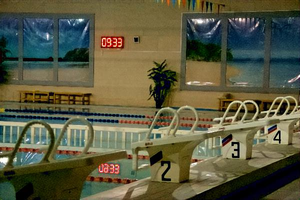,  width=\textwidth}}
		\vspace{0.1 mm}
		\centerline{\epsfig{figure=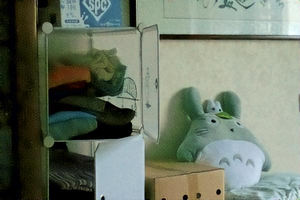,  width=\textwidth}}
		\vspace{0.1 mm}
		\centerline{\epsfig{figure=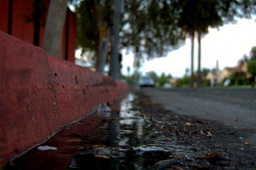,  width=\textwidth}}
		\vspace{0.1 mm}
		\centerline{\epsfig{figure=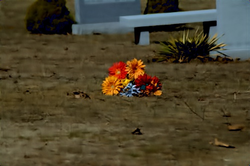,  width=\textwidth}}
		\centering{MBLLEN}
	\end{minipage}

	\begin{minipage}[b]{0.161\linewidth}
		\centering
		\centerline{\epsfig{figure=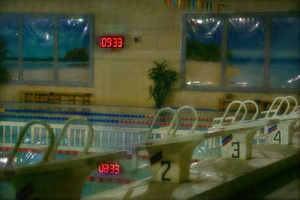,  width=\textwidth}}
		\vspace{0.1 mm}
		\centerline{\epsfig{figure=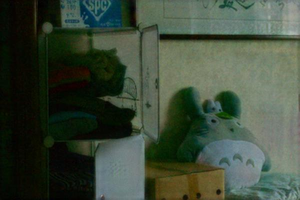,  width=\textwidth}}
		\vspace{0.1 mm}
		\centerline{\epsfig{figure=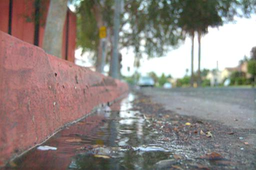,  width=\textwidth}}
		\vspace{0.1 mm}
		\centerline{\epsfig{figure=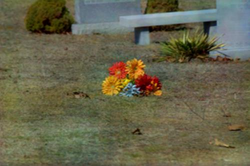,  width=\textwidth}}
		\centering{DSLR}
	\end{minipage}
	\begin{minipage}[b]{0.161\linewidth}
		\centering
		\centerline{\epsfig{figure=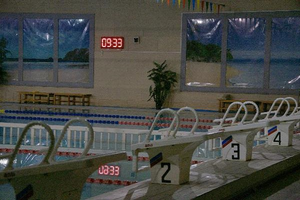,  width=\textwidth}}
		\vspace{0.1 mm}
		\centerline{\epsfig{figure=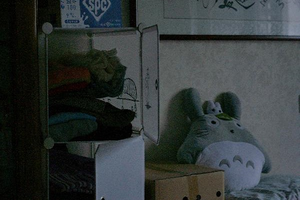,  width=\textwidth}}
		\vspace{0.1 mm}
		\centerline{\epsfig{figure=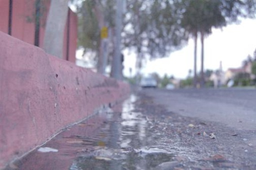,  width=\textwidth}}
		\vspace{0.1 mm}
		\centerline{\epsfig{figure=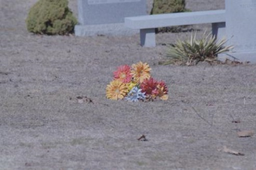,  width=\textwidth}}
		\centering{ZeroDCE++}
	\end{minipage}
	\begin{minipage}[b]{0.161\linewidth}
		\centering
		\centerline{\epsfig{figure=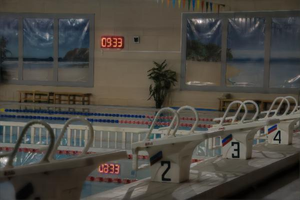,  width=\textwidth}}
		\vspace{0.1 mm}
		\centerline{\epsfig{figure=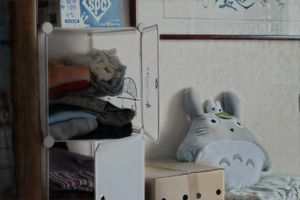,  width=\textwidth}}
		\vspace{0.1 mm}
		\centerline{\epsfig{figure=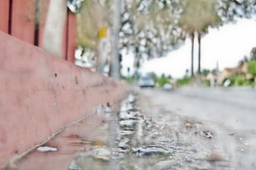,  width=\textwidth}}
		\vspace{0.1 mm}
		\centerline{\epsfig{figure=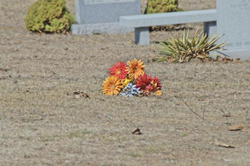,  width=\textwidth}}
		\centering{KinD++}
	\end{minipage}
	\begin{minipage}[b]{0.161\linewidth}
		\centering
		\centerline{\epsfig{figure=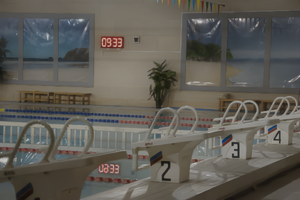,  width=\textwidth}}
		\vspace{0.1 mm}
		\centerline{\epsfig{figure=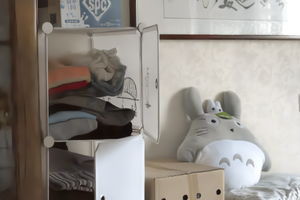,  width=\textwidth}}
		\vspace{0.1 mm}
		\centerline{\epsfig{figure=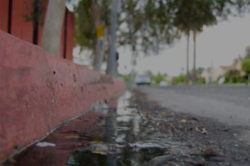,  width=\textwidth}}
		\vspace{0.1 mm}
		\centerline{\epsfig{figure=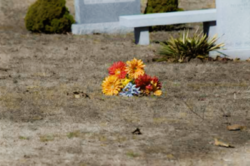,  width=\textwidth}}
		\centering{LLFlow}
	\end{minipage}
	\begin{minipage}[b]{0.161\linewidth}
		\centering
		\centerline{\epsfig{figure=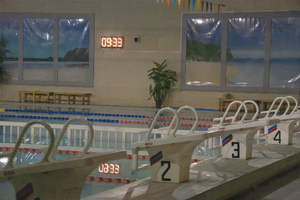,  width=\textwidth}}
		\vspace{0.1 mm}
		\centerline{\epsfig{figure=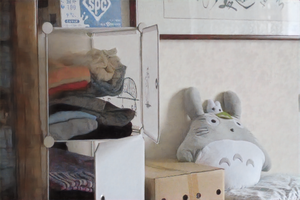,  width=\textwidth}}
		\vspace{0.1 mm}
		\centerline{\epsfig{figure=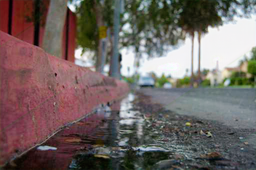,  width=\textwidth}}
		\vspace{0.1 mm}
		\centerline{\epsfig{figure=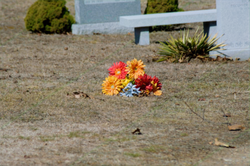,  width=\textwidth}}
		\centering{Ours}
	\end{minipage}
	\begin{minipage}[b]{0.161\linewidth}
		\centering
		\centerline{\epsfig{figure=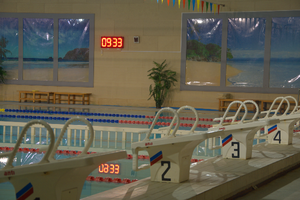,  width=\textwidth}}
		\vspace{0.1 mm}
		\centerline{\epsfig{figure=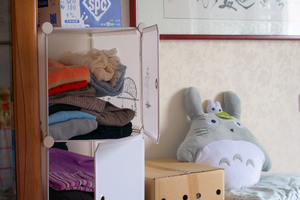,  width=\textwidth}}
		\vspace{0.1 mm}
		\centerline{\epsfig{figure=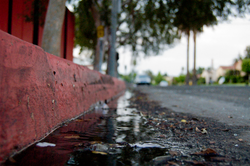,  width=\textwidth}}
		\vspace{0.1 mm}
		\centerline{\epsfig{figure=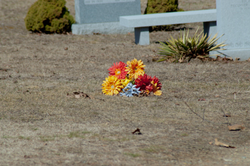,  width=\textwidth}}
		\centering{GT}
	\end{minipage}
	\caption{Visual comparisons with state-of-the-art low-light image enhancement methods on the LOL and MIT-5K datasets.}
	%Our method obtains better visual quality and recovers more image details compared with other state-of-the-art methods.
	\label{fig:lol}
\end{figure}
%----------------expriments--------------------
\section{Experiments}
\label{sec:exp}
We validate our model on two datasets, including the LOL \cite{W-Box} and the MIT-5K \cite{bychkovsky2011learning} datasets. LOL dataset contains 500 pairs of images taken by real cameras. We selected 485 pairs as the training set and 15 pairs as the test set. MIT-5K dataset contains 5000 images taken with a DSLR camera. Similar to \cite{W-Box, bychkovsky2011learning, wang2019underexposed}, we used the results of the expert C adjustment for ground-truth and used the first 4500 images for training and the last 500 for testing.
% In addition, we adopted three metrics (PSNR \cite{psnr},SSIM \cite{hore2010image} and NIQE \cite{niqe}) to quantitatively the quality of predicted results. Note that higher PSNR and SSIM or lower NIQE both indicate better results.

The implementation of the environment uses the PyTorch~\cite{pytorch} framework, trained and tested on Intel(R) Xeon(R) Gold 5218 CPU @ 2.30GHz, 128G RAM, and TITAN RTX GPU (24G RAM) platform. 
We set the batch size to 4 and trained our model using the Adam optimizer with $\beta_1$ = 0.9, $\beta_2$ = 0.999 and $\epsilon = 10^{-8}$. 
The learning rate is initialized to 0.0001, and 200 epochs are enforced, decaying by a factor of 0.5 in every 50 epochs. 
We randomly cropped the training images to 256 $\times$ 256 and normalized the pixel values to the interval [0, 1] as the input to the network.

% \subsection{Comparisons with State-of-the-art Methods}

% We select ten of the most representative and advanced low-light enhancement methods for quantitative and qualitative comparison on LOL and MIT-Adobe FiveK datasets, respectively. 
% %
% Compared methods are BIMEF \cite{ying2017bio}, CRM \cite{ren2018lecarm}, LIME \cite{guo2016lime}, RetinexNet \cite{Chen2018Retinex}, MBLLEN \cite{MELLEN}, DSLR \cite{lim2020dslr}, DRBN \cite{yang2021band}, Zero-DCE++ \cite{li2021learning}, Kind++ \cite{zhang2021beyond}, LLFlow \cite{wang2021low}. 
% %
% For the traditional algorithms (BIMEF, CRM, and LIME), we use the MATLAB code and default parameters provided by the authors to generate enhanced images. 
% %
% For data-driven based algorithms (RetinexNet, MBLLEN, DSLR, DRBN, Zero-DCE++, Kind++, and LLFlow), we retrain and test on the fixed computing device using the parameter settings and implementation details recommended by the authors to ensure fairness of comparison. 
\begin{table}[ht]
    \renewcommand\arraystretch{1.3} 
    \centering
    \caption{Compare quantization with the state-of-the-art image enhancement methods in LOL and MIT-5K datasets.}
    \resizebox{\textwidth}{!}{
    \begin{tabular}{cc|cccccccccccc}
    \toprule
        \multicolumn{2}{c|}{} & BIMEF & CRM & LIME & RetinexNet & MBLLEN & DSLR & DRBN & ZeroDCE++ & KinD++ & LLFlow & Ours \\
        \hline
        \multirow{3}{*}{LOL}   & PSNR$\uparrow$ & 13.88 & 17.20 & 16.76 & 16.77 & 17.56 & 18.24 & 20.13 & 19.43 & 21.30  & 22.92  & \textbf{24.15}      \\
        & SSIM$\uparrow$ & 0.577 & 0.644 & 0.564 & 0.567 & 0.736 & 0.787 & 0.802 & 0.768 & 0.822  & 0.837  & \textbf{0.849}      \\
        & NIQE$\downarrow$ & 7.69  & 8.02  & 9.13  & 9.73  & \textbf{3.46}  & 4.11  & 4.63  & 7.79 & 5.11 & 4.03 & 3.94      \\
        \hline
        \multirow{3}{*}{MIT5K} & PSNR$\uparrow$ & 18.67 & 13.99 & 11.21 & 20.81 & 16.42 & 17.02 & 20.95 & 16.46 & 22.01  & 25.03  & \textbf{25.33}      \\
        & SSIM$\uparrow$ & 0.693 & 0.674 & 0.667 & 0.687 & 0.851 & 0.750 & 0.794 & 0.766 & 0.832  & 8.521  & \textbf{0.910}      \\
        & NIQE$\downarrow$ & 3.87  & 4.22  & 4.50  & 4.48  & 4.19 & 3.90  & 5.44  & 3.92 & 4.15 & 3.49 & \textbf{3.42}    \\
    \bottomrule
    \end{tabular}}
\label{tab:tab1}
\end{table}
\vspace{- 2mm}

\subsection{Performance Comparison} 
\noindent\textbf{Quantitative Evaluation.}
Our proposed method is compared with ten low-light image enhancement methods \cite{ying2017bio,ren2018lecarm,guo2016lime,Chen2018Retinex,MELLEN,lim2020dslr,yang2021band,li2021learning,zhang2021beyond,wang2021low} on the LOL and MIT-5K datasets, and the results are shown in Table \ref{tab:tab1}. Among them, the results of SSIM \cite{hore2010image}, PSNR \cite{psnr}, and NIQE \cite{niqe} are the average values of the corresponding test sets, and the \textbf{bolded data} indicate the best results for that test set. As can be seen from Table \ref{tab:tab1}, our method can achieve better performance compared to other low-light image enhancement methods. On the LOL dataset, our method achieves 1.23dB improvement in PSNR and 0.012dB in SSIM compared to the previous LLFlow method. Furthermore, our method also significantly outperforms other methods on the MIT-5K dataset.

\begin{figure*}
	\centering
	\begin{minipage}[b]{0.16\linewidth}
		\centering
		\centerline{\epsfig{figure=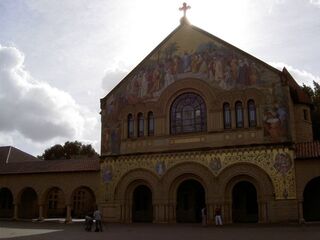,  width=\textwidth}}
		\vspace{0.1 mm}
% 		\centerline{\epsfig{figure=./imgs/nopairs/2_input.JPG,  width=\textwidth}}
% 		\vspace{0.2 mm}
% 		\centerline{\epsfig{figure=./imgs/nopairs/3_input.png,  width=\textwidth}}
% 		\vspace{0.2 mm}
		\centerline{\epsfig{figure=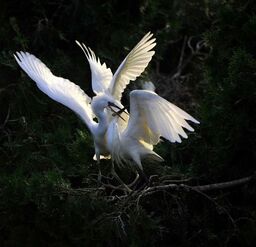,  width=\textwidth}}
		\centering{(a) Input}\medskip
	\end{minipage}
	\begin{minipage}[b]{0.16\linewidth}
		\centering
		\centerline{\epsfig{figure=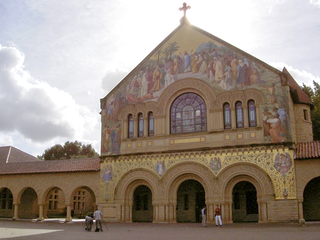,  width=\textwidth}}
		\vspace{0.1 mm}
% 		\centerline{\epsfig{figure=./imgs/nopairs/2_bimef.PNG,  width=\textwidth}}
% 		\vspace{0.2 mm}
% 		\centerline{\epsfig{figure=./imgs/nopairs/3_bimef.PNG,  width=\textwidth}}
% 		\vspace{0.2 mm}
		\centerline{\epsfig{figure=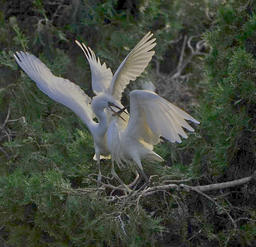,  width=\textwidth}}
		\centering{(b) BIMEF}\medskip
	\end{minipage}
	%
% 	\begin{minipage}[b]{0.07\linewidth}
% 		\centering
% 		\centerline{\epsfig{figure=./imgs/nopairs/1_crm.png,  width=\textwidth}}
% 		\vspace{0.2 mm}
% 		\centerline{\epsfig{figure=./imgs/nopairs/2_crm.png,  width=\textwidth}}
% 		\vspace{0.2 mm}
% 		\centerline{\epsfig{figure=./imgs/nopairs/3_crm.png,  width=\textwidth}}
% 		\vspace{0.2 mm}
% 		\centerline{\epsfig{figure=./imgs/nopairs/4_crm.png,  width=\textwidth}}
% 		\centering{(c) CRM}\medskip
% 	\end{minipage}
% 	\hfill
	%
	\begin{minipage}[b]{0.16\linewidth}
		\centering
		\centerline{\epsfig{figure=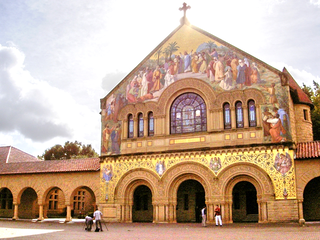,  width=\textwidth}}
		\vspace{0.1 mm}
% 		\centerline{\epsfig{figure=./imgs/nopairs/2_lime.jpg,  width=\textwidth}}
% 		\vspace{0.2 mm}
% 		\centerline{\epsfig{figure=./imgs/nopairs/3_lime.png,  width=\textwidth}}
% 		\vspace{0.2 mm}
		\centerline{\epsfig{figure=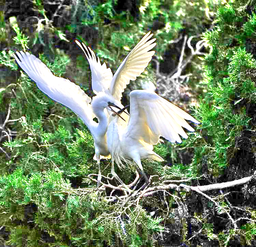,  width=\textwidth}}
		\centering{(c) LIME}\medskip
	\end{minipage}
	%
% 	\begin{minipage}[b]{0.07\linewidth}
% 		\centering
% 		\centerline{\epsfig{figure=./imgs/nopairs/1_retinex.jpg,  width=\textwidth}}
% 		\vspace{0.2 mm}
% 		\centerline{\epsfig{figure=./imgs/nopairs/2_retinex.jpg,  width=\textwidth}}
% 		\vspace{0.2 mm}
% 		\centerline{\epsfig{figure=./imgs/nopairs/3_retinex.jpg,  width=\textwidth}}
% 		\vspace{0.2 mm}
% 		\centerline{\epsfig{figure=./imgs/nopairs/4_retinex.jpg,  width=\textwidth}}
% 		\centering{(e) RetinexNet}\medskip
% 	\end{minipage}
% 	\hfill
	%
% 	\begin{minipage}[b]{0.07\linewidth}
% 		\centering
% 		\centerline{\epsfig{figure=./imgs/nopairs/1_mbllen.jpg,  width=\textwidth}}
% 		\vspace{0.2 mm}
% 		\centerline{\epsfig{figure=./imgs/nopairs/2_mbllen.jpg,  width=\textwidth}}
% 		\vspace{0.2 mm}
% 		\centerline{\epsfig{figure=./imgs/nopairs/3_mbllen.jpg,  width=\textwidth}}
% 		\vspace{0.2 mm}
% 		\centerline{\epsfig{figure=./imgs/nopairs/4_mbllen.jpg,  width=\textwidth}}
% 		\centering{(f) MBLLEN}\medskip
% 	\end{minipage}
% 	\hfill
	%
% 	\begin{minipage}[b]{0.07\linewidth}
% 		\centering
% 		\centerline{\epsfig{figure=./imgs/nopairs/1_dslr.JPG,  width=\textwidth}}
% 		\vspace{0.2 mm}
% 		\centerline{\epsfig{figure=./imgs/nopairs/2_dslr.JPG,  width=\textwidth}}
% 		\vspace{0.2 mm}
% 		\centerline{\epsfig{figure=./imgs/nopairs/3_dslr.JPG,  width=\textwidth}}
% 		\vspace{0.2 mm}
% 		\centerline{\epsfig{figure=./imgs/nopairs/4_dslr.JPG,  width=\textwidth}}
% 		\centering{(g) DSLR}\medskip
% 	\end{minipage}
% 	\hfill
	%
	\begin{minipage}[b]{0.16\linewidth}
		\centering
		\centerline{\epsfig{figure=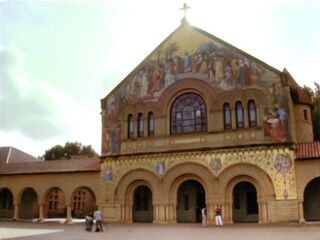,  width=\textwidth}}
		\vspace{0.1 mm}
% 		\centerline{\epsfig{figure=./imgs/nopairs/2_drbn.jpg,  width=\textwidth}}
% 		\vspace{0.2 mm}
% 		\centerline{\epsfig{figure=./imgs/nopairs/3_drbn.jpg,  width=\textwidth}}
% 		\vspace{0.2 mm}
		\centerline{\epsfig{figure=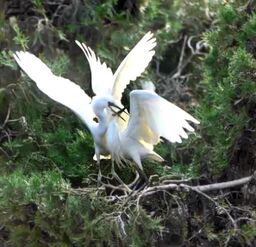,  width=\textwidth}}
		\centering{(d) DRBN}\medskip
	\end{minipage}
% 	\hfill
	%
% 	\begin{minipage}[b]{0.07\linewidth}
% 		\centering
% 		\centerline{\epsfig{figure=./imgs/nopairs/1_kind++.jpg,  width=\textwidth}}
% 		\vspace{0.2 mm}
% 		\centerline{\epsfig{figure=./imgs/nopairs/2_kind++.jpg,  width=\textwidth}}
% 		\vspace{0.2 mm}
% 		\centerline{\epsfig{figure=./imgs/nopairs/3_kind++.jpg,  width=\textwidth}}
% 		\vspace{0.2 mm}
% 		\centerline{\epsfig{figure=./imgs/nopairs/4_kind++.jpg,  width=\textwidth}}
% 		\centering{(i) KinD++}\medskip
% 	\end{minipage}
% 	\hfill
	%
% 	\begin{minipage}[b]{0.16\linewidth}
% 		\centering
% 		\centerline{\epsfig{figure=./imgs/nopairs/1_zero_dce++.JPG,  width=\textwidth}}
% 		\vspace{0.2 mm}
% 		\centerline{\epsfig{figure=./imgs/nopairs/2_zero_dce++.JPG,  width=\textwidth}}
% 		\vspace{0.2 mm}
% 		\centerline{\epsfig{figure=./imgs/nopairs/3_zero_dce++.JPG,  width=\textwidth}}
% 		\vspace{0.2 mm}
% 		\centerline{\epsfig{figure=./imgs/nopairs/4_zero_dce++.JPG,  width=\textwidth}}
% 		\centering{(j) ZeroDCE++}\medskip
% 	\end{minipage}
	%
	\begin{minipage}[b]{0.16\linewidth}
		\centering
		\centerline{\epsfig{figure=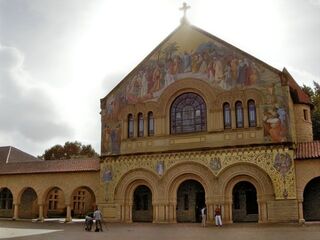,  width=\textwidth}}
		\vspace{0.1 mm}
% 		\centerline{\epsfig{figure=./imgs/nopairs/2_llflow.jpg,  width=\textwidth}}
% 		\vspace{0.2 mm}
% 		\centerline{\epsfig{figure=./imgs/nopairs/3_llflow.png,  width=\textwidth}}
% 		\vspace{0.2 mm}
		\centerline{\epsfig{figure=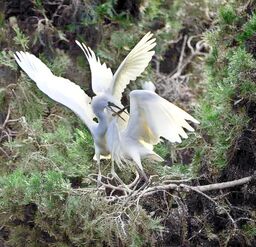,  width=\textwidth}}
		\centering{(e) LLFlow}\medskip
	\end{minipage}
	\begin{minipage}[b]{0.16\linewidth}
		\centering
		\centerline{\epsfig{figure=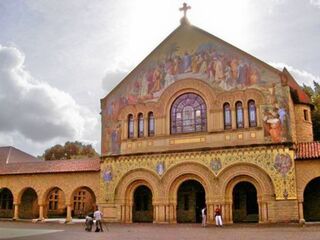,  width=\textwidth}}
		\vspace{0.1 mm}
% 		\centerline{\epsfig{figure=./imgs/nopairs/2_our.JPG,  width=\textwidth}}
% 		\vspace{0.2 mm}
% 		\centerline{\epsfig{figure=./imgs/nopairs/3_our.JPG,  width=\textwidth}}
% 		\vspace{0.2 mm}
		\centerline{\epsfig{figure=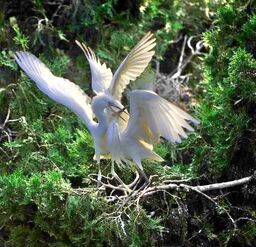,  width=\textwidth}}
		\centering{(f) Ours}\medskip
	\end{minipage}
	\caption{Visual comparisons with state-of-the-art low-light image enhancement methods on the no-reference datasets. (a) is the original input image, and (b)-(f) represent the images enhanced by SOTA and our method, respectively.}
	\label{fig:unpair}
\end{figure*}

\noindent\textbf{Qualitative Evaluation.}
In order to compare the effect of enhanced images more intuitively, we selected random images from the datasets and visualized them using the method described above. 
%
%In Figure~\ref{fig:lol}, the results show that the image visibility enhanced by the traditional physics-based methods is still insufficient and generates amplified noise. 
%
%And the images generated by the deep learning-based approach have many artifacts. 
%
According to Figure \ref{fig:lol}, we can clearly observe that the enhanced images by the traditional physics-based methods (such as BIME, CRM, and LIME) are still insufficient and generate amplified noise.
The images produced by KinD++, LLFlow, and other deep learning-based approachs still have a large number of artifacts at the edges. 
In contrast, our method enabling it to significantly enhance different degrees of darkness, and the enhanced images have more texture detail and better visual experience. In order to demonstrate the generalization ability of our network in real scenes, we selected some classical algorithms to compare the effect of enhancing real low-light images. As shown in Figure \ref{fig:unpair}, our method can equally effectively improve the real-world image of the darkness problem with an exciting performance.

%---------------- unpair image ------------------
\begin{table}[ht]
	\centering
	\caption{Quantified results of ablation experiments for network components.}
	\resizebox{0.6\textwidth}{!}{
	\begin{tabular}{ccccccccc}
		\toprule
		\multirow{2}{*}{Datasets} & \multicolumn{2}{c}{U-Net} & \multicolumn{2}{c}{MDCM} & \multicolumn{2}{c}{MDCM w/ PFM} & \multicolumn{2}{c}{DPFNet}  \\
		\cmidrule(lr){2-3}\cmidrule(lr){4-5}\cmidrule(lr){6-7}\cmidrule(lr){8-9}
		  & PSNR & SSIM & PSNR  & SSIM & PSNR & SSIM & PSNR & SSIM  \\
		\midrule
		LOL & 18.13 & 0.796 & 21.16 & 0.813 & 23.67 & 0.835 & 24.15 & 0.849  \\
		MIT5K & 20.47 & 0.825 & 23.92 & 0.851 & 24.84 & 0.874 & 25.33 & 0.910   \\
		\bottomrule
	\end{tabular}}
	\label{tab:abliation}
\end{table}
\vspace{- 2mm}
\subsection{Ablation Study}
\label{sec:ab}
%In this section, we design a series of ablation experiments to demonstrate the effectiveness of each component of the proposed method. 
% %
% First, we attempt to remove the key modules of HRGNet and analyze the enhanced images under different conditions to demonstrate the necessity and rationality of each module. 
% %
% Second, we analyze the composition of each module. 
% %
% Finally, we discuss the effectiveness of each component in the loss function.
\noindent \textbf{Effectiveness of Network Architecture.} 
We established MDCM as baseline and compared it with the classic U-Net \cite{ronneberger2015u}. Then, we sequentially add PFM (denoted MDCM w/ PFM) and AFM (DPFNet) to demonstrate the effectiveness of our proposed module. As shown in Table \ref{tab:abliation}, our proposed MDCM improves the PSNR by 3.03dB compared with UNet of frequently downsampled.
%
% We replace the adaptive fusion module and frequency domain enhancement branch by using CNN block (the kernel size is 3 $\times$ 3, a streamlined Resnet-style framework), in turn, to obtain three data with baseline (Ours), without frequency domain branch (w/o FIEB), and without adaptive module (w/o AFM). 
% We try to remove key components in the network, and sequentially get three data with baseline (HRGNet), no adaptive module (w/o AFM), and without frequency domain enhancement branch (w/o FIEB).
%
With the addition of PFM, our model improves PSNR by 2.51 dB on the LOL dataset and 0.92 dB on the MIT-5K dataset.
This also validates the importance of fusing frequency domain information with the help of PFM to enhance the enhancement effect of dark images. 
In addition, the AFM helps our network balance frequency and spatial domain features, increasing PSNR by 0.48 dB on the LOL dataset and by 0.49 dB on the MIT-5K dataset. 

\begin{figure}[ht]
% 	\begin{minipage}[b]{0.16\linewidth}
% 		\centering
% 		\centerline{\epsfig{figure=./imgs/loss/a.jpg,  width=\textwidth}}
% 		\centering{(a)}\medskip
% 	\end{minipage}
	%
	\begin{minipage}[b]{0.191\linewidth}
		\centering
		\centerline{\epsfig{figure=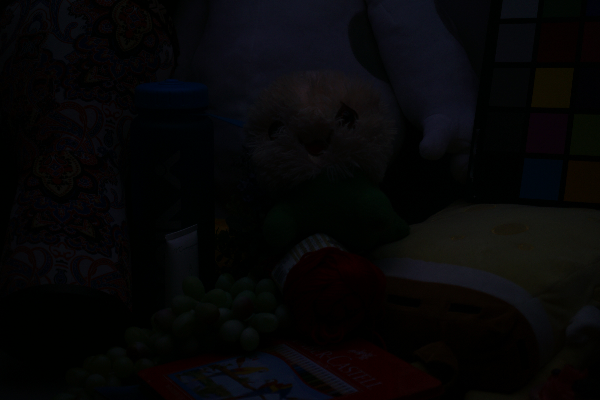,  width=\textwidth}}
		\centering{Input}
	\end{minipage}
	\begin{minipage}[b]{0.191\linewidth}
		\centering
		\centerline{\epsfig{figure=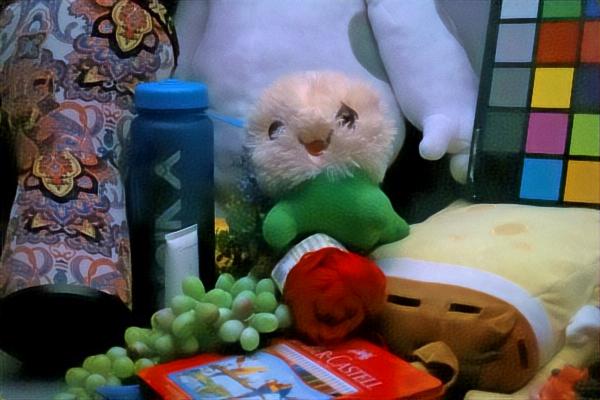,  width=\textwidth}}
		\centering{$\mathcal{L}_{s}$}
	\end{minipage}
	\begin{minipage}[b]{0.191\linewidth}
		\centering
		\centerline{\epsfig{figure=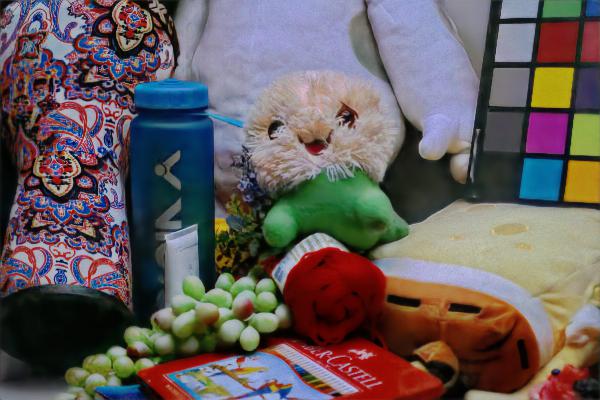,  width=\textwidth}}
		\centering{$\mathcal{L}_{s}+\mathcal{L}_{f}$}
	\end{minipage}
	\begin{minipage}[b]{0.191\linewidth}
		\centering
		\centerline{\epsfig{figure=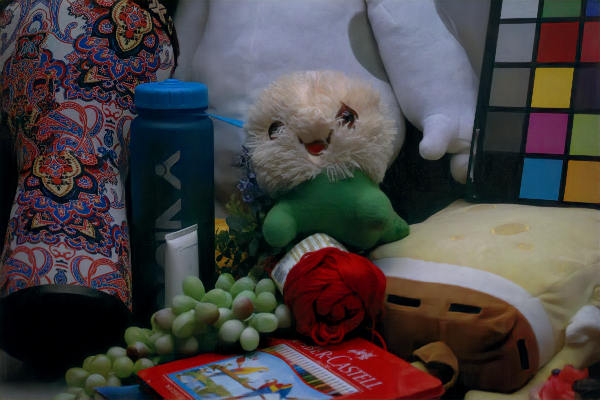,  width=\textwidth}}
		\centering{$\mathcal{L}_{s}+\mathcal{L}_{f}+\mathcal{L}_{p}$}
	\end{minipage}
	\begin{minipage}[b]{0.191\linewidth}
		\centering
		\centerline{\epsfig{figure=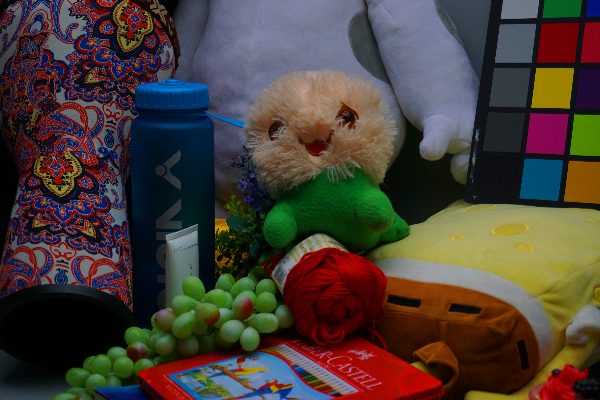,  width=\textwidth}}
		\centering{GT}
	\end{minipage}
% 	\vspace{-3mm}
	\caption{Visual results of loss component in the loss committee on the LOL dataset.}
	\label{fig:loss}
% 	\vspace{-3mm}
\end{figure}
% \vspace{-2mm}
\noindent \textbf{Effectiveness of Loss Function.} 
% In this paper, we use a loss committee consisting of SSIM loss, ReFFT loss, and VGG loss to guide the model training. 
% %
% In Table \ref{tab:loss}, we verify the importance of each component. It can be seen that the performance of the model trained using the full loss function outperforms the performance with any partial part removed. 
% %
We use the SSIM loss ($\mathcal{L}_{s}$) as the essential content loss, then the Fourier loss ($\mathcal{L}_{f}$) and Perceptual loss ($\mathcal{L}_{p}$) are added to guide the network training. Figure \ref{fig:loss} shows the intuitive comparison results, we can observe that the SSIM loss has been able to make the model produce satisfactory results, but the details are still lacking. The addition of Fourier loss helps better PFM perceive high-frequency details, but the restored image has an over-enhanced effect. Finally, it can be seen that the Perceptual loss constrains the model-enhanced images at a high level of semantics, enabling our DPFNet to produce more natural images.

\vspace{-2mm}
\section{Conclusion}
\vspace{-2 mm  }
\label{sec:con}
In this paper, we propose a novel low-light image enhancement network. For this network, we propose a Phase aware Fourier convolution module. The module can provide high-frequency information for spatial domain reconstruction by exploiting the properties of Fourier coefficients to generate sharper images. In addition, we design a dilated convolution module that aggregates features from different receptive fields to alleviate the loss caused by the downsampling of common network structures. We integrate the above parts into a two-branch framework to enhance spatial and frequency domain information. Finally, to reconstruct a more natural result, we propose a Fourier loss and adaptive fusion module to guarantee the stability of network training. Quantitative and qualitative results show that our algorithm can recover sharper images and exhibit more potential.

\bibliographystyle{splncs04}
\bibliography{mybibliography}

\begin{thebibliography}{10}
\providecommand{\url}[1]{\texttt{#1}}
\providecommand{\urlprefix}{URL }
\providecommand{\doi}[1]{https://doi.org/#1}

\bibitem{bychkovsky2011learning}
Bychkovsky, V., Paris, S., Chan, E., Durand, F.: Learning photographic global
  tonal adjustment with a database of input/output image pairs. In: CVPR 2011.
  pp. 97--104 (2011)

\bibitem{gnanasambandam2020image}
Gnanasambandam, A., Chan, S.H.: Image classification in the dark using quanta
  image sensors. In: ECCV. pp. 484--501 (2020)

\bibitem{guo2016lime}
Guo, X., Li, Y., Ling, H.: Lime: Low-light image enhancement via illumination
  map estimation. IEEE Trans Image Process  \textbf{26}(2),  982--993 (2016)

\bibitem{hore2010image}
Hore, A., Ziou, D.: Image quality metrics: Psnr vs. ssim. In: ICPR. pp.
  2366--2369 (2010)

\bibitem{hu2020dccrn}
Hu, Y., Liu, Y., Lv, S., Xing, M., Zhang, S., Fu, Y., Wu, J., Zhang, B., Xie,
  L.: Dccrn: Deep complex convolution recurrent network for phase-aware speech
  enhancement. arXiv preprint arXiv:2008.00264  (2020)

\bibitem{W-Box}
Hu, Y., He, H., Xu, C., Wang, B., Lin, S.: Exposure: A white-box photo
  post-processing framework. ACM Trans. Graph.  \textbf{37}(2),  1--17 (2018)

\bibitem{psnr}
Huynh-Thu, Q., Ghanbari, M.: Scope of validity of psnr in image/video quality
  assessment. Electron. Lett.  \textbf{44}(13),  800--801 (2008)

\bibitem{johnson2016perceptual}
Johnson, J., Alahi, A., Fei-Fei, L.: Perceptual losses for real-time style
  transfer and super-resolution. In: European conference on computer vision.
  pp. 694--711 (2016)

\bibitem{koh2022bnudc}
Koh, J., Lee, J., Yoon, S.: Bnudc: A two-branched deep neural network for
  restoring images from under-display cameras. In: CVPR. pp. 1950--1959 (2022)

\bibitem{li2021learning}
Li, C., Guo, C., Loy, C.C.: Learning to enhance low-light image via
  zero-reference deep curve estimation. arXiv preprint arXiv:2103.00860  (2021)

\bibitem{li2021deep}
Li, G., Yang, Y., Qu, X., Cao, D., Li, K.: A deep learning based image
  enhancement approach for autonomous driving at night. Knowl Based Syst
  \textbf{213},  106617 (2021)

\bibitem{liang2021swinir}
Liang, J., Cao, J., Sun, G., Zhang, K., Van~Gool, L., Timofte, R.: Swinir:
  Image restoration using swin transformer. In: Proceedings of the IEEE/CVF
  International Conference on Computer Vision. pp. 1833--1844 (2021)

\bibitem{lim2020dslr}
Lim, S., Kim, W.: Dslr: Deep stacked laplacian restorer for low-light image
  enhancement. IEEE Trans Multimedia  \textbf{23},  4272--4284 (2020)

\bibitem{liu2021inception}
Liu, J., Li, C., Liang, F., Lin, C., Sun, M., Yan, J., Ouyang, W., Xu, D.:
  Inception convolution with efficient dilation search. In: CVPR. pp.
  11486--11495 (2021)

\bibitem{llnet}
Lore, K.G., Akintayo, A., Sarkar, S.: Llnet: A deep autoencoder approach to
  natural low-light image enhancement. Pattern Recognit  \textbf{61},  650--662
  (2017)

\bibitem{MELLEN}
Lv, F., Lu, F., Wu, J., Lim, C.: Mbllen: Low-light image/video enhancement
  using cnns. In: BMVC. vol.~220 (2018)

\bibitem{mao2021deep}
Mao, X., Liu, Y., Shen, W., Li, Q., Wang, Y.: Deep residual fourier
  transformation for single image deblurring. arXiv preprint arXiv:2111.11745
  (2021)

\bibitem{niqe}
Mittal, A., Fellow, IEEE, Soundararajan, R., Bovik, A.C.: Making a 'completely
  blind' image quality analyzer. IEEE Signal Process. Lett.  \textbf{20}(3),
  209--212 (2013)

\bibitem{pytorch}
Paszke, A., Gross, S., Massa, F., Lerer, A., Bradbury, J., Chanan, G., Killeen,
  T., Lin, Z., Gimelshein, N., Antiga, L., et~al.: Pytorch: An imperative
  style, high-performance deep learning library. arXiv preprint
  arXiv:1912.01703  (2019)

\bibitem{rao2021global}
Rao, Y., Zhao, W., Zhu, Z., Lu, J., Zhou, J.: Global filter networks for image
  classification. Advances in Neural Information Processing Systems
  \textbf{34},  980--993 (2021)

\bibitem{ren2018lecarm}
Ren, Y., Ying, Z., Li, T.H., Li, G.: Lecarm: Low-light image enhancement using
  the camera response model. IEEE Trans Circuits Syst Video Technol
  \textbf{29}(4),  968--981 (2018)

\bibitem{ronneberger2015u}
Ronneberger, O., Fischer, P., Brox, T.: U-net: Convolutional networks for
  biomedical image segmentation. In: MICCAI. pp. 234--241 (2015)

\bibitem{simonyan2014very}
Simonyan, K., Zisserman, A.: Very deep convolutional networks for large-scale
  image recognition. arXiv preprint arXiv:1409.1556  (2014)

\bibitem{suvorov2022resolution}
Suvorov, R., Logacheva, E., Mashikhin, A., Remizova, A., Ashukha, A.,
  Silvestrov, A., Kong, N., Goka, H., Park, K., Lempitsky, V.:
  Resolution-robust large mask inpainting with fourier convolutions. In: WACV.
  pp. 2149--2159 (2022)

\bibitem{wang2022sfnet}
Wang, H., Chen, Y., Cai, Y., Chen, L., Li, Y., Sotelo, M.A., Li, Z.: Sfnet-n:
  An improved sfnet algorithm for semantic segmentation of low-light autonomous
  driving road scenes. IEEE trans Intell Transp Syst  (2022)

\bibitem{wang2019underexposed}
Wang, R., Zhang, Q., Fu, C.W., Shen, X., Zheng, W.S., Jia, J.: Underexposed
  photo enhancement using deep illumination estimation. In: Proceedings of the
  IEEE/CVF Conference on Computer Vision and Pattern Recognition. pp.
  6849--6857 (2019)

\bibitem{wang2021low}
Wang, Y., Wan, R., Yang, W., Li, H., Chau, L.P., Kot, A.C.: Low-light image
  enhancement with normalizing flow. In: AAAI (2022)

\bibitem{wang2022uformer}
Wang, Z., Cun, X., Bao, J., Zhou, W., Liu, J., Li, H.: Uformer: A general
  u-shaped transformer for image restoration. In: CVPR. pp. 17683--17693 (2022)

\bibitem{Retinex}
Wei, C., Wang, W., Yang, W., Liu, J.: Deep retinex decomposition for low-light
  enhancement. arXiv preprint arXiv:1808.04560  (2018)

\bibitem{Chen2018Retinex}
Wei, C., Wang, W., Yang, W., Liu, J.: Deep retinex decomposition for low-light
  enhancement. arXiv preprint arXiv:1808.04560  (2018)

\bibitem{wu2022uretinex}
Wu, W., Weng, J., Zhang, P., Wang, X., Yang, W., Jiang, J.: Uretinex-net:
  Retinex-based deep unfolding network for low-light image enhancement. In:
  CVPR. pp. 5901--5910 (2022)

\bibitem{93}
Xu, K., Qin, M., Sun, F., Wang, Y., Chen, Y.K., Ren, F.: Learning in the
  frequency domain. In: CVPR. pp. 1740--1749 (2020)

\bibitem{yang2021band}
Yang, W., Wang, S., Fang, Y., Wang, Y., Liu, J.: Band representation-based
  semi-supervised low-light image enhancement: Bridging the gap between signal
  fidelity and perceptual quality. IEEE Trans Image Process  \textbf{30},
  3461--3473 (2021)

\bibitem{ying2017bio}
Ying, Z., Li, G., Gao, W.: A bio-inspired multi-exposure fusion framework for
  low-light image enhancement. arXiv preprint arXiv:1711.00591  (2017)

\bibitem{yu2021two}
Yu, Y., Liu, H., Fu, M., Chen, J., Wang, X., Wang, K.: A two-branch neural
  network for non-homogeneous dehazing via ensemble learning. In: CVPR. pp.
  193--202 (2021)

\bibitem{zamir2022restormer}
Zamir, S.W., Arora, A., Khan, S., Hayat, M., Khan, F.S., Yang, M.H.: Restormer:
  Efficient transformer for high-resolution image restoration. In: CVPR. pp.
  5728--5739 (2022)

\bibitem{zhang2021beyond}
Zhang, Y., Guo, X., Ma, J., Liu, W., Zhang, J.: Beyond brightening low-light
  images. Int J Comput Vis  \textbf{129}(4),  1013--1037 (2021)

\bibitem{KinD}
Zhang, Y., Zhang, J., Guo, X.: Kindling the darkness: A practical low-light
  image enhancer. In: ACM MM. pp. 1632--1640 (2019)

\bibitem{zhao2016loss}
Zhao, H., Gallo, O., Frosio, I., Kautz, J.: Loss functions for image
  restoration with neural networks. IEEE Trans Comput Imaging  \textbf{3}(1),
  47--57 (2016)

\end{thebibliography}

\end{document}